\documentclass[]{interact}

\usepackage{epstopdf}
\usepackage[caption=false]{subfig}

\usepackage{multirow}

\usepackage[longnamesfirst,sort]{natbib}
\usepackage{gensymb}
\bibpunct[, ]{(}{)}{;}{a}{,}{,}
\renewcommand\bibfont{\fontsize{10}{12}\selectfont}

\theoremstyle{plain}

\theoremstyle{definition}

\theoremstyle{remark}

\begin{document}


\title{Software System for Road Condition Forecast Correction}


\author{
\name{Dmitrii Smolyakov\textsuperscript{a}\thanks{CONTACT Dmitrii Smolyakov. Email: Dmitrii.Smoliakov@skolkovotech.ru} and Evgeny Burnaev\textsuperscript{a}}
\affil{\textsuperscript{a}Skoltech, Bolshoy Boulevard 30, bld. 1, Moscow, Russia}}

\maketitle

\begin{abstract}
In this paper, we present a monitoring system that allows increasing road safety by predicting ice formation. The system consists of a network of road weather stations and intelligence data processing program module.  The results were achieved by combining physical models for forecasting road conditions based on measurements from stations and machine learning models for detecting incorrect data and forecast correction.
\end{abstract}


\section{Introduction}
Road safety is a complex issue that includes many aspects. Some of them depend on human behavior; others depend on infrastructure conditions. For example, in many regions of the Russian Federation, spring and autumn temperatures can fluctuate near 0\degree C, which in combination with rains and high humidity could lead to ice formation on the roads. Detection of these conditions is essential for road safety.

Online monitoring allows preventing road accidents by early maintenance; for example, we can use video monitoring of roads' conditions. Machine learning techniques enable detecting ice formation automatically. For instance, in~\cite{zhao2017road}, the authors used data collected from stationary cameras to distinguish between five different conditions of the road pavement. In~\cite{roychowdhury2018machine}, authors proposed to collect video from vehicle front cameras instead of static cameras, and to use a more accurate classification technique based on convolutional neural networks.

Despite all their advantages, video monitoring with ordinary cameras struggles with black ice detection. In order to fix this problem we can use multispectral cameras \cite{gregoris2004multispectral}, NIR cameras~\cite{jonsson2014road}, or measure depth with Kinect sensor \cite{abdalla2017black}. These techniques are more expensive than ordinary cameras; however, they can provide better results.

A video monitoring system is costly. It can be reasonable to use sensors that collect other characteristics of road conditions like temperature, humidity, pressure, and others. Stations collecting this sort of data are called road weather information systems (RWIS). They can be located on the sides of roads or even inside the pavement~\cite{tabatabai2017novel}.

By processing historical information on road conditions, we can predict ice formation. We can group forecasting algorithms into two major categories. The first category uses physical models~\cite{barber1957calculation, sass1997numerical, feng2012numerical,dai2003common} or their combinations like in~\cite{chunlei2009summer}. The authors combine results of the physical models Common Land Model (CoLM)  and BJ-RUC \cite{fan2009performance}, which allowed increasing the accuracy. Later in this paper, we take a closer look at the physical model METRo. It performs exceptionally well in countries like Czech Republic~\cite{sokol2014first}, Canada~\cite{crevier2001metro}, USA~\cite{rutz2013integration}, China~\cite{fan2009performance}.

Despite impressive track records, physical models have several problems. The main disadvantage is their computational complexity since they require the numerical solution of partial derivative equations.  Another challenge is the deterministic nature of their forecasts. These algorithms provide forecast values but struggle with delivering a level of confidence. It is possible to solve this issue with the Mont-Carlo simulation \cite{chapman2012probabilistic, berrocal2010probabilistic}, but it significantly increases computational complexity. Another challenge for this kind of models is anthropogenic factors, which are extremely difficult to estimate. There are some approaches with an additional prior assumption about anthropogenic factor behaviour~\cite{khalifa2016accounting}, but still, they cannot deal with stochastic effects and increase already significant computational complexity.

Another big group of techniques uses machine learning and statistical approaches to road conditions forecast. Even a simple linear regression allows to get reasonable forecasting accuracy; however, usually, people use these methods as part of more complex solutions~\cite{diefenderfer2003development, yang2010atmospheric, liao2009temperature}. Using more sophisticated machine learning algorithms like deep neural networks allows getting even better forecast accuracy~\cite{zaytar2016sequence,gensler2016deep}. However, this family of methods is not a silver bullet and has its problems. For example, a lack of physical assumptions in modeling leads to a lack of interpretability and a constant need for additional refitting to work in a new environment.

It seems that both physical and machine-learning-based techniques have their problems; this is why we decide to build a hybrid system that would combine advantages of both types of models. Another challenge is the quality of the data in such big systems of sensors; it is pretty standard that some devices become malfunctioning, and it is crucial to detect these broken sensors as soon as possible.

In this paper, we describe algorithmic techniques and software system, which allows achieving the stated goal. To build an intelligence system for analyzing RWIS data, we create a software system that takes the data from the sensors, validate if this data correct. For the validated data, a forecasting model provides forecast of road conditions, and these predictions are used to decide on further maintenance actions. We concentrate on the correction of weather forecasting using machine learning and anomaly (outlier) detection. Also, we briefly describe the implementation of the software system.

In section \ref{chapter:forecasting}, we discuss an approach for weather forecasting. We propose an algorithm that combines classical energy balance based model and correction of residuals with Gradient Boosting on Decision Trees. In the section~\ref{chapter:anomaly_detection} we present an approach for detecting broken RWIS sensors. In section~\ref{chapter:results}, we demonstrate achieved performance on real-world data collected from RWIS. And finally, in section~\ref{chapter:software}, we discuss some technical details about software system implementation.

\section{Road Weather Condition Forecasting}\label{chapter:forecasting}
In this section, we present a new approach for road weather condition forecasting based on a combination of an energy balance model and machine learning. Both of them have their pros and cons. So a combination of these techniques allows increasing the accuracy of forecasting.

\subsection{METRo}
We choose METRo (Model of the Environment and Temperature of Road) as a primary weather forecasting model~\cite{crevier2001metro}. This approach already demonstrated great performance in various cases around the globe. Initially, this model was tested on Canadian roads; climate similarity between Canada and Russia makes this model a good candidate for the final approach.

The core idea behind the model is to decompose temperature weave in several parts
\[
    R = (1 - \alpha)S + \varepsilon I - \varepsilon \sigma T^4_s - H - L_{a}E \pm L_{f}P + A,
\]
here $(1 - \alpha)S$ is an absorbed incoming radiation flux, $\varepsilon I$ is the absorbed incoming infrared radiation flux, $\sigma T^4$ is an emitted flux,  $H$ is a turbulence flux, $L_{a}E$ is the latent heat flux, $L_{f}P$ is a water phase changing flux, finally $A$ is an anthropogenetic flux.

We skip the detailed explanation of calculating these parts and coefficients selection, see the original paper for details~\cite{crevier2001metro}. For us, the anthropogenic part is the most interesting. A partial derivative equation cannot model this component because it is highly non-deterministic and so it introduces additional forecasting error. However, we can fix the error produced by this factor by using a machine learning algorithm.

\subsection{Forecast improvement with gradient boosting}

To correct the METRo forecast, we predict the difference between its result and the real values of road parameters.  Here we illustrate this approach using four parameters predicted by the METRo model: air, road, and under road temperatures and humidity. 

Let $y_t$ be one of the target variables at time $t$, $y_t^{METRo}$ be the value predicted by the METRo model. We build an approximation of the difference between the real value and predictions $\Delta y_t = y_t - y_t^{METRo}$.

We construct an approximation function $\Delta y_t \approx F(x_t)$, where $x_t$ is a feature vector that represents road condition at time $t$. These features can be grouped into three classes:
\begin{enumerate}
    \item  Historical features collected from Road Weather Information Systems. They include information about air, road, and underground temperatures, etc.
    \item Information about year season and time of the day. Since these features have a cyclic nature, we encode them by using trigonometric functions. Number of days since the beginning of the year $d$ produces two values $\sin(d/365)$ and $\cos(d/365)$ or $\sin(d/366)$ and $\cos(d/366)$ in case of a leap year. Hour of the day $h$ is transformed into pair $\sin(h/24)$ and $\cos(h/24)$.
    \item Previous predictions from METRo equations.
\end{enumerate}

We build an approximation function $F(x)$ by minimizing a specified loss function $L(\cdot,\cdot)$, which characterizes forecast accuracy:
\[
    \hat{F} = \arg\min_{F} \sum_{t=1}^TL(F(x_t), \Delta y_t).
\]

To find this optimal $\hat{F}$, we use the Gradient Boosting algorithm \cite{friedman2002stochastic}. Its main idea is to build a sequence of regression models, a combination of which provides a good approximation. At step $N$, the approximation has the form:
\[
    \hat{F}_N(x) = \sum_{i=1}^N \alpha_i h_i(x),
\]
where $h_i$ is a regression model with weight $\alpha_i$. On step $N+1$ we solve the optimization subproblem to get the model $h_{N+1}$:
\[
    Q_{N+1} = \sum_{t=1}^T L(\hat{F}_N(x_t) + \alpha_{N+1}h_{N+1}(x_t), \Delta y_t) \to \min_{\alpha_{N+1}, h_{N+1}}.
\]
To get $h_{N+1}$ we solve the subproblem:
\[
    h_{N+1} = \arg\min\limits_h\sum_{t=1}^T((-\nabla Q^t_{N+1}) - h(x_t))^2,
    \quad
    [\nabla Q^t_{N+1}]^T_{t=1} = \left[\frac{\partial L(z,\Delta y_t)}{\partial z}\Bigl.\Bigr|_{z = \hat{F}_{N}(x_t)}\right]^T_{t=1}.
\]

To find optimal $\alpha_{N+1}$ value, we solve the following optimization problem by the linear search:
\[
    \alpha_{N+1} = \arg\min\limits_{\alpha} \sum_{t=1}^TL(\hat{F}_{N}(x_t) + \alpha h_{N+1}(x_t),\Delta y_t).
\]
As a loss function we use the Mean Absolute Error:
\[
    MAE = \frac{1}{T}\sum_{t=1}^T |\Delta y_t - \hat{F}(x_t)|.
\]

For constructed $\hat{F}(x)$ the final prediction is:
\[
    \hat{y}_t = y_t^{\text{METRo}} + \hat{F}(x_t).
\]

We construct 12 different regression models to predict air, road, underground temperatures, and humidity for the forecast horizons of 1, 2, and 3 hours.

\section{Broken Sensors Detection}\label{chapter:anomaly_detection}
Accuracy of road condition forecasting depends on the quality of the data. It is crucial to detect malfunction sensors as soon as possible. Wrong data could lead either to conducting maintenance actions for no reason or, even worse, to miss the dangerous road conditions. 

Malfunctioning sensors generate abnormal data, and so we can detect broken sensors by detecting anomalies in data. By anomaly, we consider the observation that deviates from others so much that we could suspend that it came from another source \cite{hawkins1980identification}. To find these values, we build an anomaly detector, i.e., the function such that:
\[
    f(y_t) = \begin{cases}
    \phantom{-}1 & \text{if } y_t \text{ is normal,}\\
    -1 & \text{if } y_t \text{ is anomaly}.
    \end{cases}
\]



To construct the anomaly detector we
\begin{enumerate}
    \item predict a future value $\hat{y}_t$ of the target variable,
    \item compare it with the actual value $y_t$ of the target variable.
\end{enumerate}
If the difference $\Delta_t = |y_t - \hat{y}_t|$ is big, this could mean that something drastically has changed, and this could be a symptom of a broken sensor. The anomaly detection function takes the form
\begin{equation}
\label{deltaeq}
    f(y_t) = \begin{cases}
    \phantom{-}1 & \text{if }\Delta_t \leq \Delta,\\
    -1 &  \text{if } \Delta_t > \Delta.\\
    \end{cases}
\end{equation}
Accurate selection of threshold $\Delta$ is very important: we have to reliably distinguish between cases of just random fluctuations and anomalies due to  malfunctioning sensors. 

\begin{figure}[t]
    \centering
    \includegraphics[width=0.9\linewidth]{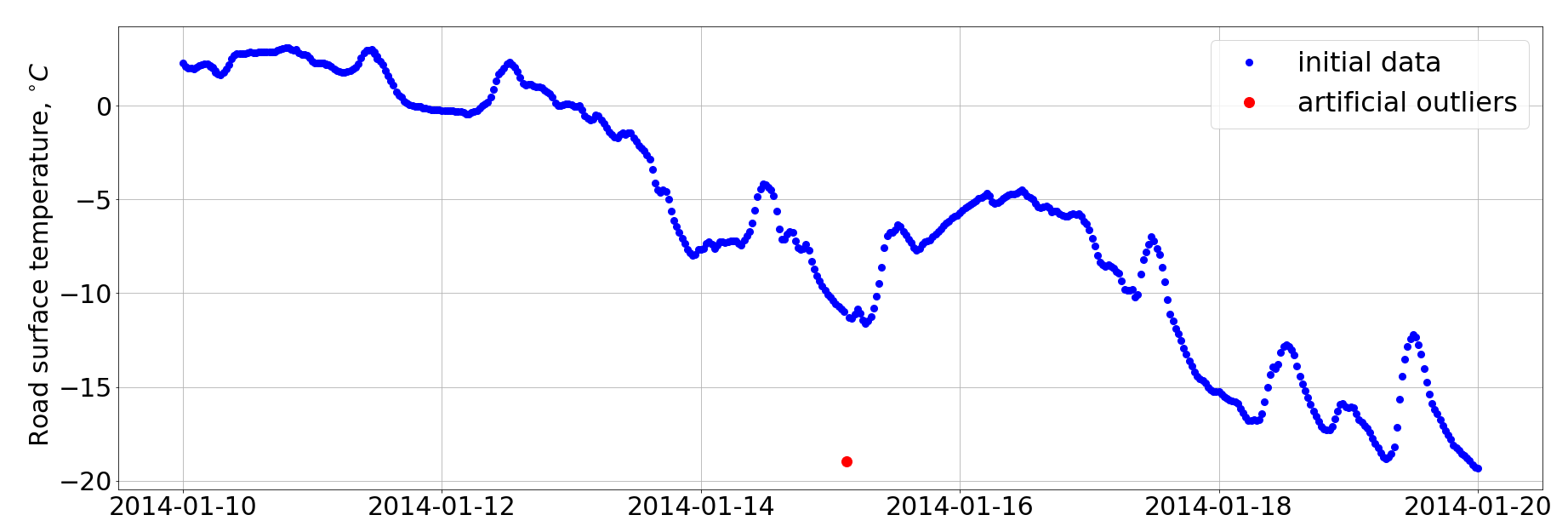}
    \caption{Single Anomaly}
    \label{fig:single_outlier}
\end{figure}

\begin{figure}[t]
    \centering
    \includegraphics[width=0.9\linewidth]{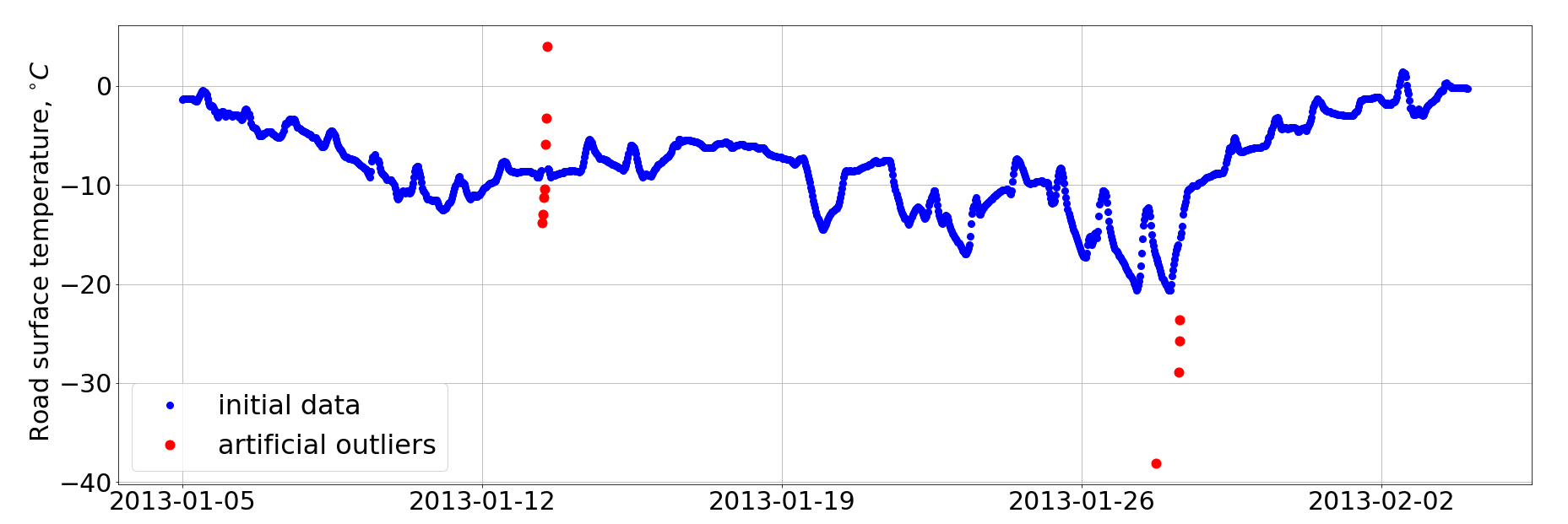}
    \caption{Short Term Anomaly}
    \label{fig:short_outlier}
\end{figure}

\begin{figure}[t]
    \centering
    \includegraphics[width=0.9\linewidth]{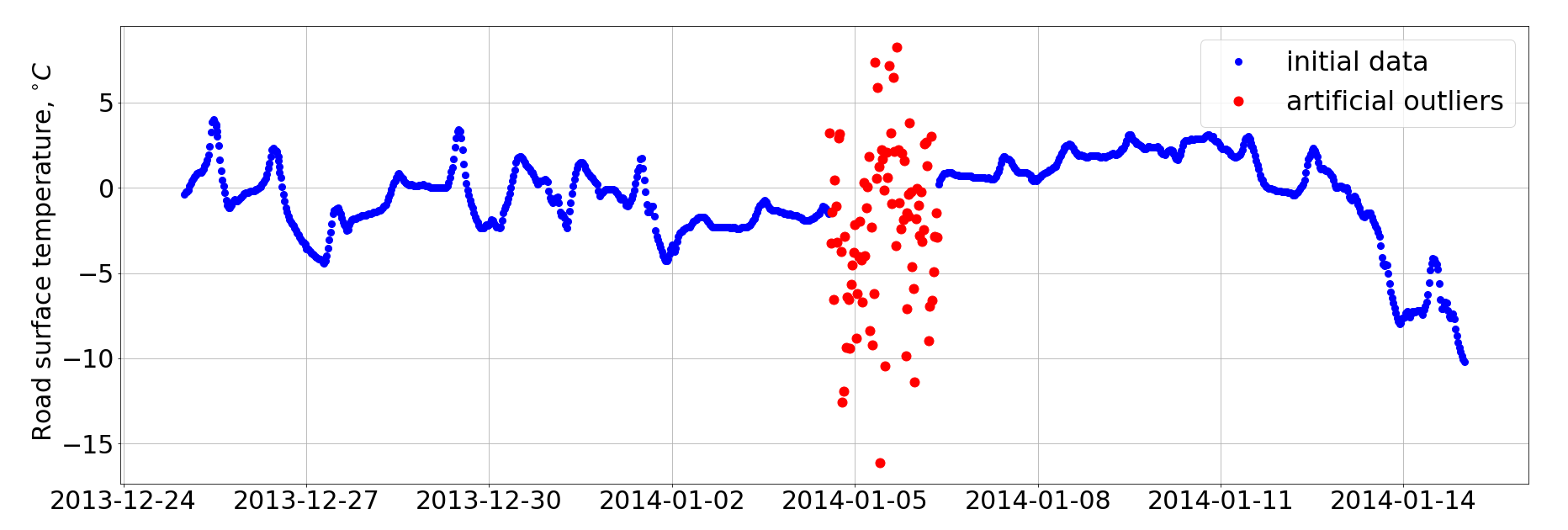}
    \caption{Long Term Anomaly}
    \label{fig:long_outlier}
\end{figure}

 Usually, the value of $\Delta$ is selected based on a holdout set or using the cross-validation procedure~\cite{friedman2001elements}. However, for that we need labeled data. Since broken sensors are  not frequent events, even such a significant amount of data like we have does not contain enough examples of anomalies. Moreover, labeling is a tedious task. Therefore, we propose to use artificially generated anomalies to select the threshold. We group all anomalies into three different categories:
\begin{enumerate}
    \item Single anomalies, which can result in some deviation. They are modeled with uniform distribution, see Fig. \ref{fig:single_outlier}.
    \item Short-term anomalies modeled as a Poisson noise with random parameters see Fig. \ref{fig:short_outlier}.
    \item  Long-term anomalies, which can be results of some significant dis-functioning in sensors or networks. They were modeled with a Gaussian noise, see Fig. \ref{fig:long_outlier}.
\end{enumerate}

To find threshold $\Delta$, we select a subset of data, add these artificial anomalies, and find the value of $\Delta$, which provides the optimal $F_1$ score:
\begin{equation}\label{eq:f1}
    F_1 = \frac{\text{precision} \cdot \text{recall}}{\text{precision}+ \text{recall}},
\end{equation}
where recall is the fraction of correctly identified anomalies, while precision is the fraction of anomalies we are able to find.

\section{Results}\label{chapter:results}
\subsection{Forecasting Correction}
This section discusses the experimental results of road condition forecasting based on a combination of the METRo model and gradient boosting residual correction. 

We collected data for seven years from 2012 till 2019 from 120 RWIS, located in different regions of Russia. This data was split into two parts. We fit the model based on data for the period 2012 -- 2017 years. We also split data based on different RWIS locations. So, in the end, we have the test data from a later period of time, and from RWIS that were not used while building a model. We measured the Mean Average Error between the predicted value and the real one.

\begin{figure}[t]
    \centering
    \includegraphics[width=0.9\linewidth]{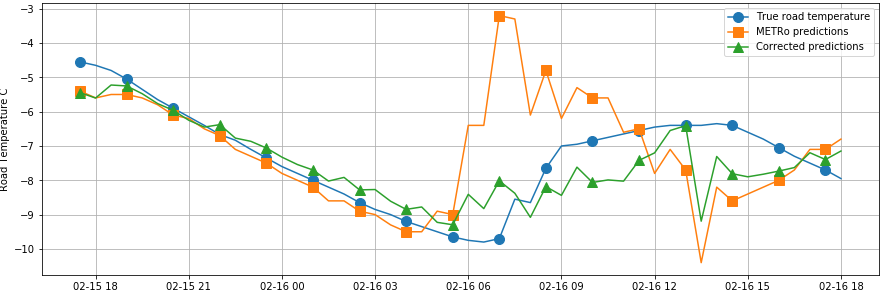}
    \caption{Road Temperature Forecasting Example}
    \label{fig:predictions}
\end{figure}

We compare our results with and without additional residual corrections. Example of prediction are presented in Fig.~\ref{fig:predictions}.  Table \ref{tab:forecasting} demonstrates that additional machine learning post-processing allows for increasing the accuracy of the final forecast.

\begin{table}[t]
    \centering
    \caption{Forecasting Improvements}
    \label{tab:forecasting}
    \begin{tabular}{|c|c|c|c|c|}
    \hline
    Value   &   Model   &   1 hour  &   2 hours     &   3 hours  \\
    \hline
     \multirow{3}{*}{Air Temperature \degree C}  & \multicolumn{1}{l|}{METRo with anomalies} & \multicolumn{1}{l|}{0.69} & \multicolumn{1}{l|}{1.01} & \multicolumn{1}{l|}{1.25} \\
     &\multicolumn{1}{l|}{METRo only} & \multicolumn{1}{l|}{0.63} &\multicolumn{1}{l|}{0.97} &\multicolumn{1}{l|}{1.18}\\
     &\multicolumn{1}{l|}{METRo + Correction} & \multicolumn{1}{l|}{\textbf{0.53}} &\multicolumn{1}{l|}{\textbf{0.78}} &\multicolumn{1}{l|}{\textbf{0.93}}\\
    \hline
    \multirow{3}{*}{Road Temperature \degree C} & \multicolumn{1}{l|}{METRo with anomalies} & \multicolumn{1}{l|}{0.93} & \multicolumn{1}{l|}{1.69} & \multicolumn{1}{l|}{2.8} \\
     & \multicolumn{1}{l|}{METRo only} & \multicolumn{1}{l|}{0.84} &\multicolumn{1}{l|}{1.66} &\multicolumn{1}{l|}{2.41}\\
        & \multicolumn{1}{l|}{METRo + Correction} & \multicolumn{1}{l|}{\textbf{0.48}} &\multicolumn{1}{l|}{\textbf{0.89}} &\multicolumn{1}{l|}{\textbf{1.28}}\\
    \hline
     \multirow{3}{*}{Underground Temperature \degree C} &\multicolumn{1}{l|}{METRo with anomalies} & \multicolumn{1}{l|}{1.45} & \multicolumn{1}{l|}{1.7} & \multicolumn{1}{l|}{1.94} \\
    &\multicolumn{1}{l|}{METRo only} & \multicolumn{1}{l|}{1.39} &\multicolumn{1}{l|}{1.62} &\multicolumn{1}{l|}{1.81}\\
        &\multicolumn{1}{l|}{METRo + Correction} & \multicolumn{1}{l|}{\textbf{1.16}} &\multicolumn{1}{l|}{\textbf{1.48}} &\multicolumn{1}{l|}{\textbf{1.79}}\\
    \hline
    \end{tabular}
\end{table}

\subsection{Anomaly Detection}
This section  discusses experimental results of predicting broken sensors of RWIS. Let us consider experimental settings, metrics and final accuracy indicators.

We select 59 RWIS stations from various regions of the Russian Federation. We use data from 50 stations for training models and data from 9 stations for testing the quality of the resulting model. The training set was split into two parts. Data from 35 RWIS was used to build a model; data from another 15 RWIS was used to tune threshold value based on artificially generated anomalies.

We labeled data (normal/anomaly) from the 9 test stations manually. Final results were evaluated by using $F_1$ score.

We compare our approach with other anomaly detection techniques like Elliptic Envelope~\cite{butler1993asymptotics}, Isolation Forrest~\cite{liu2008isolation}, OneClassSVM with Gaussian kernel~\cite{scholkopf2000new}. We also test different regression algorithms to construct approximation $\hat{F}(x)$ beside gradient boosting over regression trees, specifically ridge regression~\cite{marquardt1975ridge} and multi-layer perceptron~\cite{murtagh1991multilayer}. Results of the comparison are presented in Table~\ref{table:anomaly}. We can see that the proposed approach based on gradient boosting outperforms all other algorithms in terms of the $F_1$ score, see formula~\ref{eq:f1}.  Table \ref{tab:forecasting} demonstrates that if before making predictions with METRo model we remove anomalous data, we can significantly improve forecasting accuracy. Note that when detecting anomalies we did not use METRo model for estimating $\Delta_t$ in formula \ref{deltaeq}.

\begin{table}[t]
\centering
    \caption{Anomaly Detection Results}
    \label{table:anomaly}
    \begin{tabular}{|c|c|c|c|c|c|c|c|}
    \hline
         Algorithm   & LOF   & Ell.Env & OCSVM  & IForest & Ridge & MLP   & Boosting\\
        \hline
         Recall      & 0.513 & 0.719   & 0.904  & 0.395   & 0.668 & 0.637 & 0.753     \\
        \hline
        Precision    & 0.502 & 0.482   & 0.130  & 0.252   & 0.758 & 0.738 & 0.797     \\
        \hline
        $F_1$-score  & 0.507 & 0.577   & 0.227  & 0.308   & 0.710 & 0.684 & 0.774   \\
        \hline
    \end{tabular}
    
    \end{table}

\section{Implementation}\label{chapter:software}

Initially, company MinMax94\footnote{https://mm94.ru} has already installed a system for road condition forecasting, which consists of the METRo module and decision-making module connected to the MongoDB. We developed a system that could be easily added to this existed infrastructure.

To do this, we developed two web services based on python library Flask and put them into a docker container. They communicate with the outer world by HTTP requests.

\begin{figure}[t]
    \centering
    \includegraphics[width=0.9\linewidth]{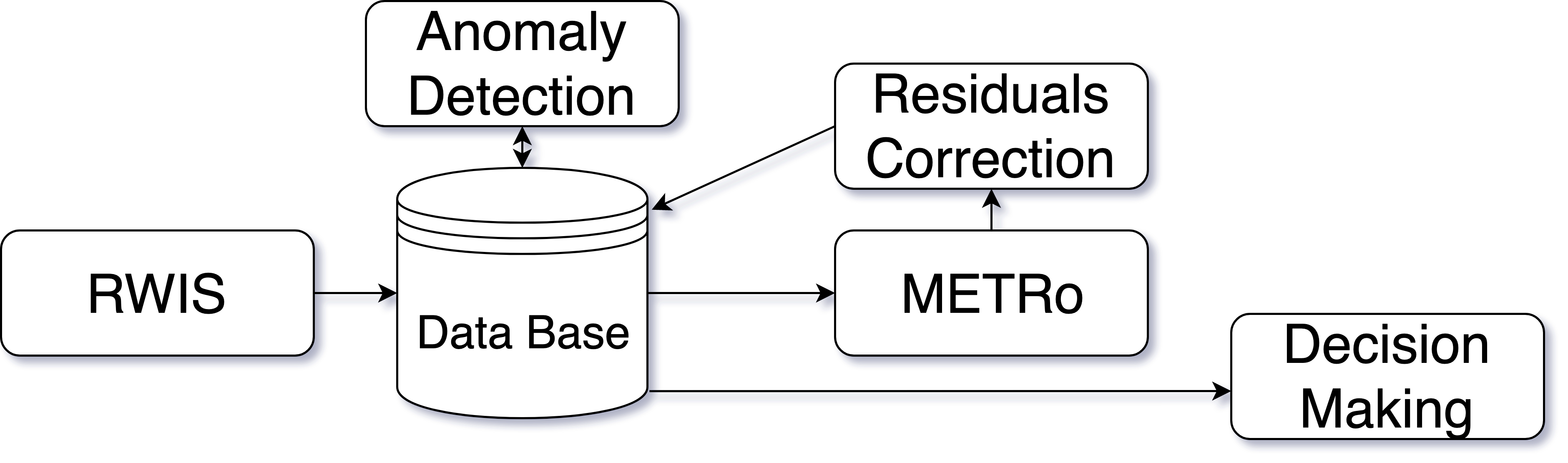}
    \caption{Data Flow}
    \label{fig:data_flow}
\end{figure}

The dataflow is presented in Fig.~\ref{fig:data_flow}. So the whole work process can be descrived as follows: data from RWIS goes into the database, then the data is sent to the Anomaly Detection Module. The module generates labels (normal/anomaly) for each observation. If there are many anomalies, we don't use information from these sensors before they would be checked and fixed. Information about which observations are anomalous is saved in the database.

If the data is normal we send it to the METRo module, which builds predictions. The results are sent to the Residuals Correction module. Results of this module are sent back to the database, and finally, they are passed into the decision-making module, which decides if the road needs maintenance or not.

Models for Anomaly Detection and Residuals Corrections are constructed in advance. We collect historical data in CSV format, as described in section~\ref{chapter:results}. Then we use the lightgbm library~\cite{ke2017lightgbm} to build these predictive models. The whole process takes around 30 minutes on a desktop computer with the Intel Core i7 processor and 32Gb RAM.

\section{Conclusion}

We proposed a novel approach to improve road surface condition forecasting. It allows for combining the advantages of both energy-balanced and machine learning-based models. Numerical results demonstrate that despite the very impressive accuracy of the METRo model, it is still possible to improve accuracy by correcting residuals of predictions.
We describe details of the implementation of these algorithms as a part of the software system for road condition forecasting.

\bibliography{interactapasample}

\end{document}